\documentclass[11pt]{article}

\usepackage[preprint]{acl}

\usepackage{times}
\usepackage{latexsym}

\usepackage[T1]{fontenc}

\usepackage[utf8]{inputenc}

\usepackage{microtype}

\usepackage{inconsolata}

\usepackage{graphicx}
\usepackage{amsmath}
\usepackage{array}
\usepackage{caption}
\usepackage{comment}

\title{Value-Aware Numerical Representations for Transformer Language Models}

\author{Andreea Dutulescu \and Stefan Ruseti \and Mihai Dascalu \\
         National University of Science and Technology POLITEHNICA Bucharest \\ \texttt{\{andreea.dutulescu, stefan.ruseti, mihai.dascalu\}@upb.ro}}

\begin{document}
\maketitle
\begin{abstract}
Transformer-based language models often achieve strong results on mathematical reasoning benchmarks while remaining fragile on basic numerical understanding and arithmetic operations. A central limitation is that numbers are processed as symbolic tokens whose embeddings do not explicitly encode numerical value, leading to systematic errors. We introduce a value-aware numerical representation that augments standard tokenized inputs with a dedicated prefix token whose embedding is explicitly conditioned on the underlying numerical value. This mechanism injects magnitude information directly into the model’s input space while remaining compatible with existing tokenizers and decoder-only Transformer architectures. Evaluation on arithmetic tasks shows that the proposed approach outperforms baselines across numerical formats, tasks, and operand lengths. These results indicate that explicitly encoding numerical value is an effective and efficient way to improve fundamental numerical robustness in language models.
\end{abstract}

\section{Introduction}

Large Language Models (LLMs) based on the Transformer architecture have shown remarkable capabilities across a wide range of natural language processing tasks, including question answering, code generation, and complex multi-step reasoning. Recent models achieve competitive performance on challenging mathematical benchmarks and standardized examinations, often by leveraging large-scale pretraining, instruction tuning \citep{cobbe2021gsm8k, hendrycks2math, toshniwal2025openmathinstruct}, and explicit reasoning-enhancing training \citep{shao2024deepseekmath}. These advances have led to the widespread perception that LLMs possess strong mathematical and numerical abilities.

However, growing empirical evidence shows that this perception is incomplete. Despite their apparent success on complex reasoning problems, LLMs frequently fail at elementary numerical understanding and basic arithmetic operations. As highlighted by \citet{yang2024numbercookbook}, modern LLMs can produce convincing reasoning traces while making errors, such as incorrect numerical comparisons (e.g., concluding that 9.11 > 9.9) or simple arithmetic mistakes involving fractions. These failures indicate that numerical competence is not a natural by-product of improved reasoning, but a distinct capability that remains fragile in current architectures. Many benchmarks conflate high-level reasoning (e.g., problem interpretation, equation formulation, or procedural planning) with low-level numerical processing. As argued by \citet{yang2024numbercookbook}, a typical Math problem requires both reasoning about its structure and correctly manipulating the numbers involved; yet evaluation datasets rarely disentangle these components. Consequently, improvements attributed to "better reasoning" may mask persistent weaknesses in basic numerical understanding.

\begin{figure}[t]
    \centering
    \includegraphics[width=\linewidth]{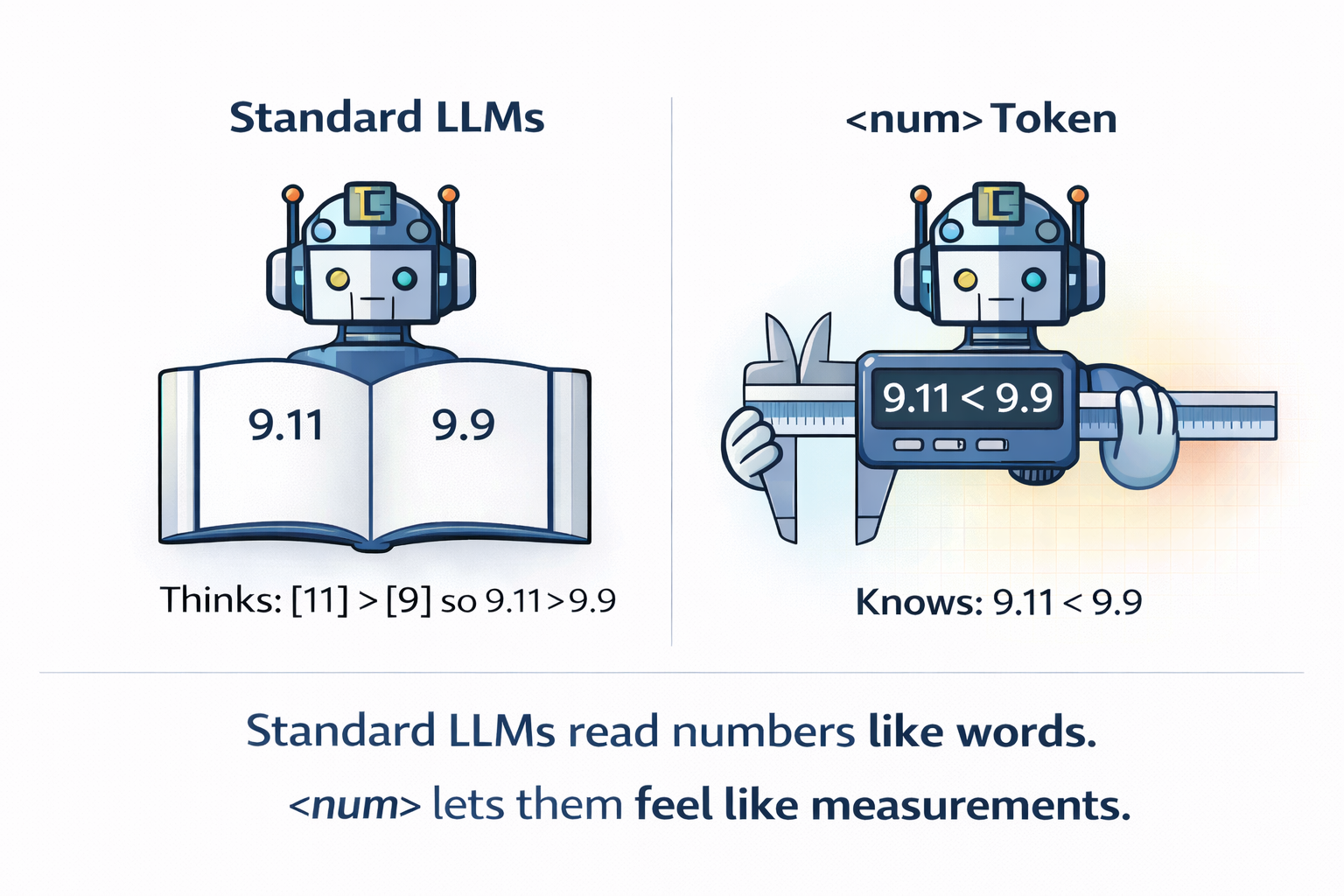}
    \caption{Standard LLMs process numbers as symbolic tokens, which can lead to incorrect surface-form arithmetics.
    Our value-aware \texttt{<num>} token explicitly encodes numerical magnitude, enabling the model to reason over numbers as continuous measurements.}
    \label{fig:teaser_pic}
\end{figure}

A dominant trend in recent research focuses on enhancing Math capabilities through methods that generate long intermediate reasoning chains or agentic workflows. While these approaches often improve final-answer accuracy, they come at the cost of generating many additional tokens and significantly increasing inference time. More importantly, they do not address a fundamental limitation: even when guided step-by-step, models may still lack a reliable internal representation of numerical value. In other words, producing more reasoning tokens does not resolve the fact that models often do not "know" what a number represents.

At the core of this problem lies the notion of numerical value. For humans, a number has an intrinsic magnitude independent of its surface form. For Transformer-based LLMs, however, numbers are primarily treated as singular or sequences of tokens, and their semantics emerge indirectly from distributional patterns in text. Models process numbers piece by piece rather than as unified quantities, leading to systematic errors in comparison, magnitude estimation, and arithmetic as the input length grows.

In this paper, we propose a complementary approach to numerical representation in Transformers. Instead of encoding numbers solely through their surface tokens, we introduce a value-aware prefix token that precedes each numerical expression. The embedding of this token is explicitly conditioned on its numerical value, providing the model with a continuous, magnitude-sensitive signal alongside the standard tokenized representation. This design preserves compatibility with existing tokenizers and architectures while injecting inductive bias about numerical values directly into the input representation.

Our main contributions are as follows:
\begin{itemize}
    \item We propose a value-aware numerical encoding mechanism for language models, in which numerical expressions are augmented with a dedicated prefix token whose embedding is explicitly conditioned on the underlying numerical value, rather than being learned solely from token co-occurrence statistics.
    
    \item We empirically confirm that our approach consistently improves on a benchmark designed to isolate arithmetic competence, outperforming strong baselines under similar training and inference conditions.
    
    \item Our proposed mechanism requires only minimal architectural modifications and is directly applicable to any decoder-only Transformer architecture. To support reproducibility and adoption, we release the full implementation, including architectural changes, training code, and evaluation scripts (\url{https://anonymous.4open.science/r/Temp-Value-Aware-Numeric-Repr-8796/}). 
\end{itemize}

By decoupling numerical value from surface tokenization while retaining the flexibility of standard language modeling, our approach targets a fundamental gap identified by prior work: the absence of an explicit notion of number magnitude in LLM representations. Encoding numbers using our approach provides a lightweight, efficient alternative to reasoning-intensive methods, and directly addresses the root causes of basic numerical errors rather than compensating for them through extended inference.

\section{Related Work}
\subsection{Improving Mathematical Performance via Reasoning and Diverse Methods}

A substantial body of recent work improves mathematical performance in LLMs by modifying the reasoning process itself, rather than directly targeting numerical representations or arithmetic operations. These approaches typically assume that errors arise from insufficient or poorly structured reasoning and therefore focus on eliciting, refining, or controlling intermediate solution steps.

\citet{didolkar2024metacognitive} introduced a framework in which models first identified the skills required to solve a mathematical problem and then condition generation on skill-specific in-context examples. \citet{wang2025formalreason, chen2025predicate} used modules that translate the Math problem into formal language, which is then offered as additional input for solving. Other approaches \citep{cao2025step, leang2025comat} relied on different forms and prompts for step-by-step reasoning to enhance Math capabilities. Additionally, to improve Math reasoning capabilities, multiple synthetic datasets and data generation methods have been proposed \citep{toshniwal2025openmathinstruct, wan2025empowering, xiao2025ab}.

Another direction examined whether explicit step-by-step reasoning is always necessary. \citet{liu2024skipsteps} studied whether models can be trained to produce shorter or partially implicit reasoning traces without sacrificing correctness. By fine-tuning on datasets containing both detailed and abbreviated rationales, the authors showed that models can learn to omit intermediate steps while maintaining or even improving accuracy. This work highlighted that reasoning verbosity and reasoning quality are not strictly coupled.

Finally, several approaches \citealp{wang2022selfconsistency, lee2025revise} focused on test-time verification and refinement. \citet{lee2025revise} proposed a self-verification mechanism that allowed models to evaluate and revise their own reasoning trajectories during inference. Through a curriculum-based training procedure and confidence-aware decoding, models learnt to detect potential errors and refine intermediate reasoning steps without external verifiers. 

Overall, these methods achieve gains by strengthening reasoning control and self-correction. However, they largely treat arithmetic correctness as a downstream consequence of improved reasoning processes, rather than as a capability that can be independently modeled and enhanced at the level of numerical representation or value understanding.

\subsection{Improving Arithmetics}

Relatively few studies focus on arithmetic performance in isolation, decoupled from higher-level reasoning. Recent work has nevertheless proposed targeted interventions at many levels, aiming to improve numerical accuracy independently of extended reasoning strategies.

Standard numeral representations do not explicitly expose digit length or place value, requiring models to infer scale implicitly during left-to-right generation. \citet{schwartz2024numerologic} addressed this issue by introducing a structured number format in which each numeral is prefixed with its digit count. This representation provides early access to magnitude information and reduces ambiguity in multi-digit arithmetic. The approach does not modify model architectures or training procedures and can be applied at inference time. Empirical results show improvements on arithmetic benchmarks and general evaluation tasks with numerical content.

A complementary direction targets the training objective used for number prediction. Standard cross-entropy loss treats number tokens as discrete, unrelated classes and therefore ignores numerical proximity between predictions. \citet{zausinger2025regress} proposed a loss function that incorporated regression-style signals into language model training. This encouraged predictions that are numerically closer to the target even when token-level accuracy is not exact. The method can be integrated into existing training pipelines and has been shown to improve arithmetic accuracy relative to models trained with cross-entropy alone.

\citet{cheng2025disentangling} separated abstract problem formulation from arithmetic computation in mathematical reasoning tasks and showed that, despite its higher conceptual complexity, abstract formulation is handled more effectively by large language models than numerical computation. Their results further indicated that Chain-of-Thought prompting yields only marginal improvements in arithmetic accuracy.

\citet{dugan2024occamllm} introduced OccamLLM, a framework in which an LLM’s hidden states are mapped to a symbolic computation module that performs arithmetic deterministically. Unlike approaches based on multi-step code generation or external tool invocation, OccamLLM executes arithmetic in a single step and does not require fine-tuning the base language model. This design yielded exact results on arithmetic operations and argued that symbolic computation can be tightly integrated with neural language models to guarantee correctness on numerical tasks.

Overall, these approaches improve numerical performance by modifying surface objectives or execution pathways, but they do not introduce an explicit representation of numerical value within the model’s internal processing. As a result, these methods mitigate symptoms of numerical failure without directly addressing the lack of a value-aware representation in Transformer-based LLMs.

\subsection{Tokenization}

Most language models rely on subword tokenization schemes such as Byte Pair Encoding \citep{sennrich2016bpe} or SentencePiece \citep{kudo2018sentencepiece}, which are optimized for corpus compression rather than semantic structure. While effective for natural language, these approaches are ill-suited to numerals, whose meaning is determined by digit identity and place value \citep{yang2024numbercookbook}. 

To address this issue, different model families have adopted distinct tokenization strategies for numbers. Palm \citep{chowdhery2023palm}, early versions of the Llama family \citep{touvron2023llama2}, as well as Qwen models \citep{yang2025qwen3}, enforced single-digit tokenization, ensuring that each digit corresponds to a unique token. In contrast, the latest open-source GPT models \citep{agarwal2025gptoss} and newer Llama versions \citep{dubey2024llama3} define individual tokens for all 1-, 2-, and 3-digit numbers. While this reduces sequence length and improves inference efficiency, it introduces a large numeric vocabulary and nontrivial inductive biases.

Recent empirical studies show that these tokenization choices substantially impact numerical reasoning. \citet{singh2024tokenizationcounts} highlighted that left-to-right 3-digit chunking, as used by GPT-style tokenizers, undermines arithmetic accuracy, and that reversing the chunking direction or enforcing digit-level splits improves performance. Controlled experiments found that single-digit tokenization consistently yielded better arithmetic performance \citep{yang2024numbercookbook}.

\section{Method}
\begin{figure*}[!ht]
    \centering
    \includegraphics[width=\linewidth]{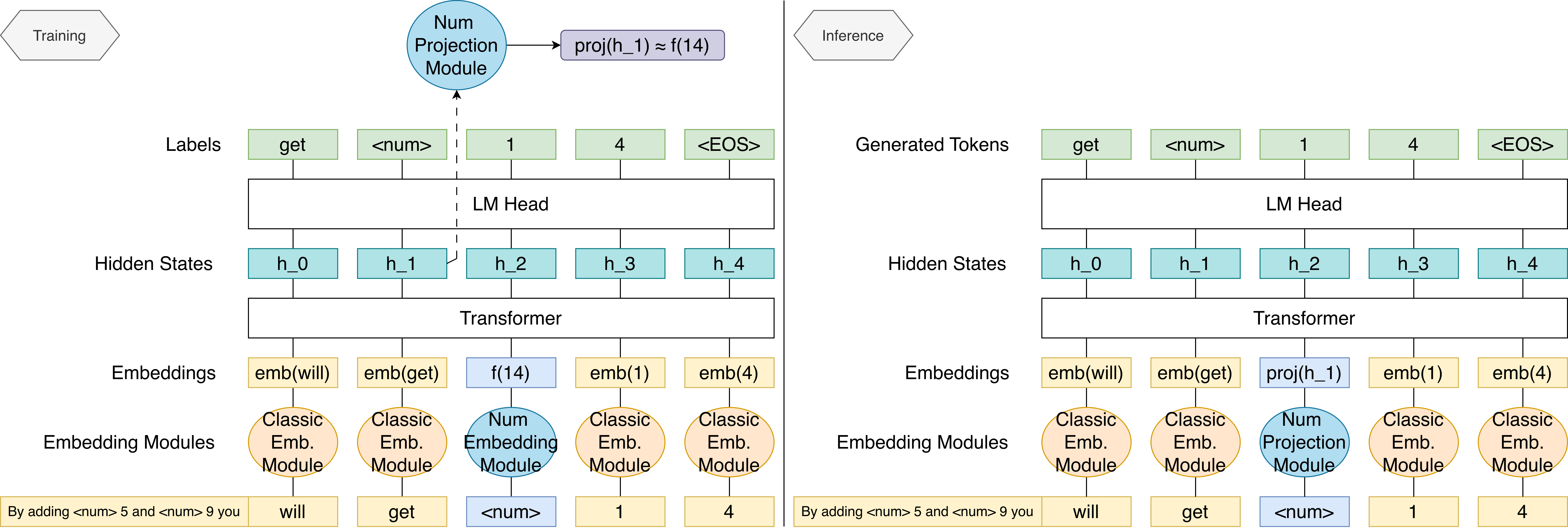}
    \caption{Training and inference overview of the proposed approach. A standard Transformer is augmented with a special \texttt{<num>} token inserted before surface-form number tokens to explicitly encode numerical value. During training (left), the true numeric value $x$ is used by a trainable embedding module to compute the \texttt{<num>} embedding as $f(x)$, and a projection head is trained to align the hidden state with this representation, enforcing $proj(h_1) \approx f(x)$. During inference (right), when $x$ is unavailable, the projected hidden state $proj(h_1)$ replaces $f(x)$ and conditions the generation of numeric tokens.}
    \label{fig:training_inference_overview}
\end{figure*}

\subsection{Preamble}
\paragraph{Motivation.} Transformer-based models do not encode numerical magnitude explicitly. Numbers are processed as tokens or token sequences whose embeddings are learned independently of the values they denote, and numerical semantics emerge only indirectly from distributional patterns in text. This representation is insufficient for capturing basic magnitude relationships and leads to systematic errors in comparison, ordering, and arithmetic, particularly as numerical length increases. Context alone provides weak supervision for learning numerical value: many numbers occur too infrequently during pretraining, while others are over-represented and dominate the embedding space (e.g., how many times does the token \texttt{153} need to appear in pretraining so that the model can "know" how it is different from \texttt{154}?). As a result, embedding similarity does not reliably correspond to numerical proximity.

\paragraph{Idea.} We explicitly encode numerical values by introducing a special \texttt{<num>} token whose embedding is not fixed, but computed as a function of the numeric value that follows. Formally, the embedding of <num> is defined as $emb(<num>) = f(x)$, where $x$ is the numerical value and $f$ is a learnable function. This token is inserted immediately before the surface-form number tokens (e.g., "they have \texttt{<num>} 14 apples"), allowing the model to access magnitude and value information while preserving the original tokenization.

\subsection{Architecture Overview}

The proposed architecture (see Figure \ref{fig:training_inference_overview}) augments a standard Transformer with value-aware numerical processing while preserving compatibility with existing tokenization and generation mechanisms. Numbers are introduced into the input sequence by prepending a special \texttt{<num>} token before the surface-form numeric tokens (e.g., "[Together] [they] [have] [\texttt{<num>}] [1][4] [apple][s]"). During training, the embedding of \texttt{<num>} is computed as $emb(<num>) = f(x)$, where $x$ is the numerical value (e.g., 14) and $f$ is a learnable function. This embedding is injected into the model alongside the standard token embeddings. 

During decoding, when the model predicts \texttt{<num>}, the numerical value is not directly available; instead, the current hidden state $h$ is projected into the embedding space via a learned projection $proj(h)$. This projection replaces $f(x)$ and is used to condition the generation of subsequent numeric tokens (e.g., [1][4]) and other future tokens. To align training and inference behavior, the projection is trained to approximate the true value-based embedding, enforcing $proj(h) \approx f(x)$.


\paragraph{\texttt{<num>} Embedding Module.}

Several designs for the value encoder $f$ were considered: (1) direct encoding of the scalar value through a fully connected network, which is simple but unstable for large magnitudes and insensitive to fine-grained fractional differences; (2) fixed-point binary or decimal representations, which improve numerical structure but suffer from limited range and uneven parameter updates; (3) or recurrent encoders over decimal representations, using separate RNNs for integer and fractional parts, which enable arbitrary precision and extensibility.

Additional auxiliary features, such as sign, digit length, and number of decimal places, can be concatenated to enrich the representation.

We experimented with two approaches for the number embedding module: fixed-point decimal representation and recurrent encoding over decimal digits. In both settings, we also concatenated auxiliary features, including sign, number of digits, and number of decimal places. 

In the first setting (MLP), the integer and fractional parts of a number are represented as fixed-length decimal vectors, right-padded to predefined maximum sizes. This composite representation is then projected through a feed-forward layer followed by a normalization-compatible activation function.

In the second setting (RNN), the integer and fractional parts are processed independently using Recurrent Neural Networks (i.e., a Gated Recurrent Unit - GRU), allowing the model to naturally handle variable-length representations and arbitrary numerical precision. GRUs model long-term dependencies with fewer parameters and lower computational cost than other RNNs. The final hidden states of the two recurrent encoders are concatenated with the same auxiliary features and mapped to the model hidden dimension via a linear projection. This design preserves flexibility with respect to number magnitude and precision.

\begin{figure}[t]
    \centering
    \includegraphics[width=\linewidth]{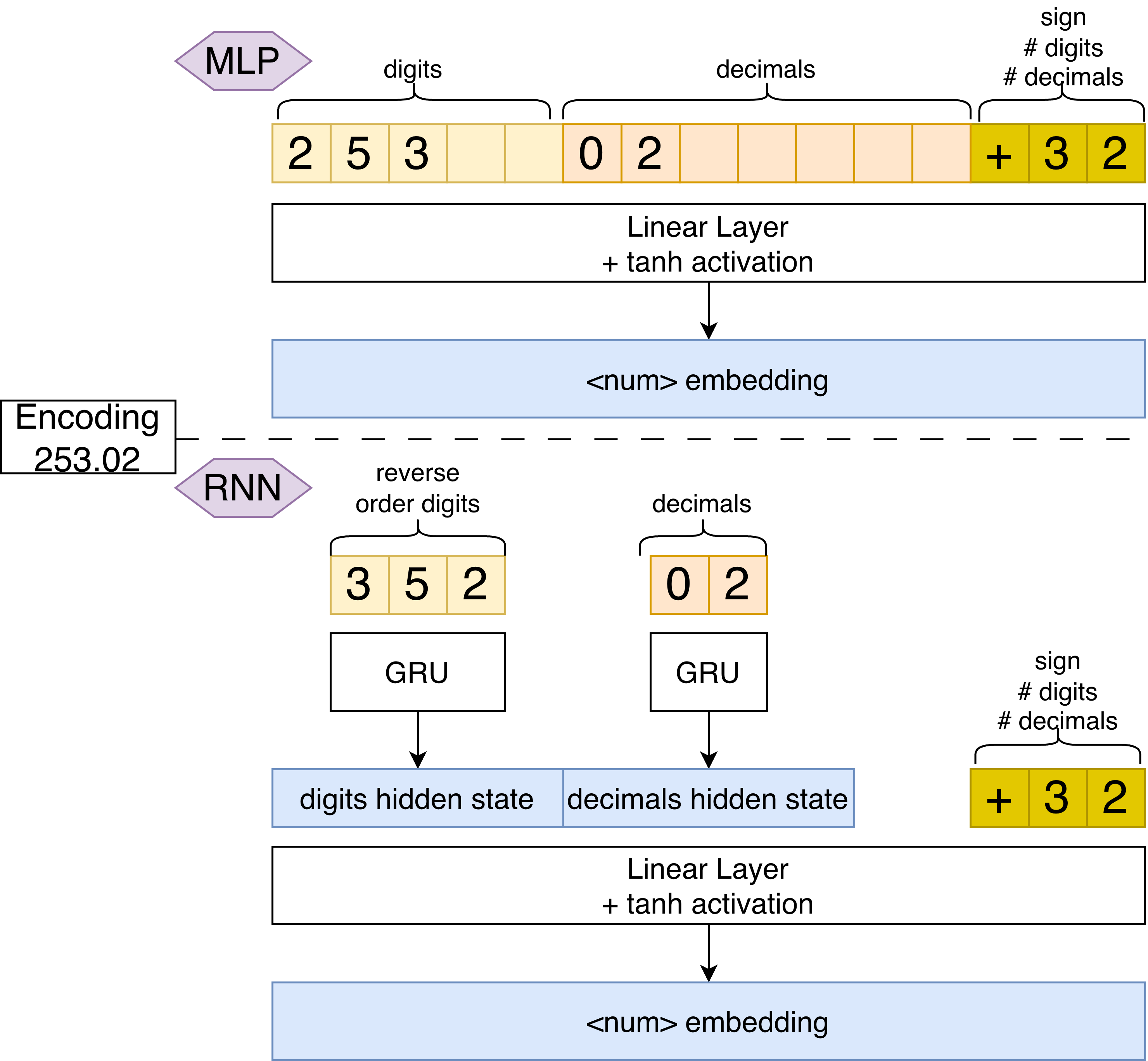}
    \caption{\texttt{<num>} embedding module variants. The MLP-based encoder represents integer and fractional parts as fixed-length decimal vectors projected through a feed-forward layer, while the RNN-based encoder processes variable-length digit sequences using separate GRUs for integer and fractional components. Auxiliary features (i.e., sign, digit count, decimal length) are concatenated in both cases.}
    \label{fig:num_embedding}
\end{figure}

An overview of these different setups for the \texttt{<num>} embedding module is shown in Figure \ref{fig:num_embedding}.

\paragraph{Training Objective.}
Training is performed by minimizing a composite loss with three complementary components that explicitly account for the dual role of the \texttt{<num>} token. Let $y = (y_1, \dots, y_T)$ denote the target token sequence and let $\mathcal{L}_{\text{CE}}(\cdot)$ be the standard cross-entropy loss for next-token prediction. Let $f(x)$ denote the value-based embedding of a numerical value $x$, and let $proj(h_t)$ denote the projection of the hidden state preceding the \texttt{<num>} token.

\begin{enumerate}
    \item The first loss corresponds to standard teacher-forced language modeling, where the embedding of \texttt{<num>} is given by the true value encoder, thus providing oracle access to the numerical value:
    
    $ \mathcal{L}_{\text{LM}}^{\text{emb}} = \mathcal{L}_{\text{CE}}\big(p_\theta(y \mid \texttt{<num>} \mapsto f(x)) \big)$
    
    \item To simulate inference-time behavior, we replace the value-based embedding of \texttt{<num>} with the projected hidden state. Let $\tilde{e}_t$ denote the input embedding sequence such that
    $\tilde{e}_t =
    \begin{cases}
    proj(h_{t-1}) & \text{if } y_t = \texttt{<num>} \\
    e_t           & \text{otherwise}
    \end{cases}
    $

    The corresponding language modeling loss is then defined as:
    
    $\mathcal{L}_{\text{LM}}^{\text{proj}} = \mathcal{L}_{\text{CE}} \big (p_\theta(y \mid \texttt{<num>} \mapsto proj(h)) \big)$

    \item To explicitly align the projected hidden state with the true value-based embedding and reduce the discrepancy between training and inference representations, we introduce a reconstruction loss defined as the cosine distance between the two representations:
    
    $\mathcal{L}_{\text{rec}} = 1 - cos\_sim\big( proj(h_{t-1}), f(x) \big)$
\end{enumerate}

The overall training objective is a weighted sum of the three loss components:
\[
\boxed{
\mathcal{L}
= \mathcal{L}_{\text{LM}}^{\text{emb}}
+ \mathcal{L}_{\text{LM}}^{\text{proj}}
+ \lambda \, \mathcal{L}_{\text{rec}}
}
\]
where $\lambda$ is a scalar hyperparameter controlling the contribution of the reconstruction loss.

The motivation for a 3-component loss arose from a train–test mismatch: during standard training, the model can rely on the true \texttt{<num>} embedding $f(x)$, effectively "cheating" when generating numeric tokens, whereas at inference time it must rely on $proj(h)$. Empirically, even with reconstruction loss, the discrepancy between these embeddings can degrade generation quality. To address this, we also replaced the true value-based embedding with a projected embedding during training, ensuring consistent behavior between training and inference. This requires two forward passes (i.e., one to compute the projection and one to decode using it), thus doubling training cost.

\section{Performance Evaluation}
\subsection{Benchmark}
\label{sec:benchmark}
We rely on the NUPA benchmark introduced by \citet{yang2024numbercookbook} to evaluate numerical understanding and processing capabilities for our approach and baselines. NUPA was specifically designed to isolate and measure fundamental numerical skills independently of higher-level linguistic reasoning or complex problem interpretation. It offers train, validation, and test partitions.

The benchmark systematically covers four numerical representations (i.e., integers, floating-point numbers, fractions, and scientific notation) and evaluates them across a wide variety of elementary tasks grouped into four categories: elementary arithmetic, comparison, digit-level understanding, and representation conversion. NUPA controls task difficulty through explicit variation in number length, enabling the analysis of model robustness with respect to input scale. In addition to exact-match accuracy, the benchmark introduces auxiliary metrics such as digit-level accuracy and length deviation. Example of training and testing samples can be found in Appendix \ref{sec:app_dataset_example_tasks}.

In our experiments, we use a subset of NUPA tasks targeting number value understanding, including all numerical representations, elementary arithmetic, comparison, and representation conversion, while excluding digit-level manipulation. This selection is intended to evaluate how models represent and reason over numerical quantities, rather than their capability to process symbolic digit sequences. For efficiency and reproducibility, we limit input operands to at most 7 digits, yielding results of up to 14 digits depending on the operation. This setting is sufficient to reveal non-trivial numerical errors while keeping computational cost and evaluation latency low. The exact task subset, number of samples, and data constraints are detailed in Appendix \ref{sec:app_dataset}.

\subsection{Baselines}

As a primary baseline, we use a standard Transformer-based language model, trained and evaluated in a conventional autoregressive setup, where numerical values are represented solely as text tokens and the model is required to predict the target number directly. This setting reflects the default numerical processing behavior of contemporary language models and serves as a reference point for assessing improvements.

In addition, we adopt Numerologic \citep{schwartz2024numerologic} as an additional baseline. Its authors proposed it as a strong approach to numerical reasoning; however, the empirical evaluation was conducted on a minimal set of tasks involving numbers with only a few digits (up to 4). In this setup, the model is augmented with explicit magnitude-related supervision expressed in natural language. Specifically, before the numerical prediction target, the model is trained to generate auxiliary textual descriptions indicating the number of digits and decimal places of the output. This method provides coarse-grained numerical structure cues while maintaining a purely text-based input–output interface.


\subsection{Experimental Setup}
We use a uniform experimental setup across all models, including our proposed approach and the two baselines, to ensure comparability of results. All models share the experimental setup described here, differing only in the modeling components under evaluation.

We adopt GPT-2 \citep{radford2019gpt2} as the base architecture for all experiments. Training is performed from scratch using the NUPA \citep{yang2024numbercookbook} training set, restricted to the task subset and numerical ranges described in Section \ref{sec:benchmark}. Model hyperparameters and optimization details are reported in the Appendix \ref{sec:app_hyperparameters}.

Training follows a curriculum learning strategy \citep{soviany2022curriculum} to progressively increase numerical complexity. Models are first trained on tasks involving operands of up to 3 digits. The training set is then expanded to include tasks with operands of up to 5 digits, while retaining a smaller proportion of 3-digit examples. Finally, models are trained on tasks with operands of up to 7 digits, again maintaining a reduced subset of up to 5-digit examples. This curriculum is designed to stabilize training.

For tokenization, we employ a digit-level tokenizer, motivated by the findings of \citet{yang2024numbercookbook} who indicated that digit-based tokenization is more effective for numerical understanding tasks. In addition, we construct a task-specific tokenizer with a reduced vocabulary, omitting non-numerical tokens, to improve training efficiency while preserving full dataset coverage.

\section{Results}
We report results on the test split of the NUPA benchmark \citep{yang2024numbercookbook}, following the same evaluation protocol as in the original benchmark paper. Evaluation is performed on the subset of tasks used during training, with operands restricted to numerical values with up to 7 digits, consistent with the training configuration.

The NUPA benchmark defines one primary evaluation metric, \emph{Exact Match}, and two auxiliary metrics: \emph{Digits Match}, which measures digit-level overlap between prediction and ground truth, and \emph{Distance in Length} (\texttt{-dlength}), which captures discrepancies in the number of digits. These complementary metrics provide a more fine-grained view of numerical prediction quality beyond strict exactness.

Table~\ref{tab:aggregated_results} summarizes overall performance by reporting averages across all evaluated tasks and operand magnitudes. We report Exact Match, Digit Match, and digit-length error (d-Length) to capture complementary aspects of numerical prediction quality, including strict correctness, partial digit-level accuracy, and structural deviation in output length.

\begin{table*}[ht]
\centering
\captionsetup{justification=centering}
\begin{tabular}{l||c||c|c}
    \hline
    \textbf{Method} & \textbf{Exact Match $\uparrow$} & \textbf{Digit Match $\uparrow$} & \textbf{d-Length $\downarrow$}\\
    \hline
    Standard Transformer & 0.687 & 0.839 & 0.068\\
    Numerologic \citep{schwartz2024numerologic} & 0.633 & 0.781 & 1.039\\
    NumValue-MLP (ours) & 0.720 & \underline{0.864} & \underline{0.060}\\
    NumValue-RNN (ours) & \textbf{0.724} & 0.862 & 0.090\\
    \hline
\end{tabular}
\caption{Aggregated results across tasks and magnitude.\\$\uparrow$ denotes a higher value is better, $\downarrow$ denotes a lower value is better.}
\label{tab:aggregated_results}
\end{table*}

To analyze the impact of operand magnitude on model behavior, Table~\ref{tab:exact_match_digits} reports Exact Match accuracy stratified by the number of digits in the operands. This breakdown highlights how performance changes with increasing numerical length and allows for a direct comparison of robustness across models.

\begin{table*}[t]
\centering
\small
\begin{tabular}{r|cccc}
\hline
\textbf{\# Digits} &
\textbf{Standard Transformer} &
\textbf{Numerologic} &
\textbf{NumValue-MLP (ours)} &
\textbf{NumValue-RNN (ours)} \\
\hline
1 & 0.719 & \textbf{0.837} & 0.833 & 0.814 \\
2 & 0.657 & \textbf{0.780} & 0.684 & 0.677 \\
3 & 0.733 & 0.675 & \textbf{0.776} & \textbf{0.776} \\
4 & 0.604 & 0.610 & 0.656 & \textbf{0.663} \\
5 & 0.646 & 0.599 & 0.681 & \textbf{0.689} \\
6 & 0.619 & 0.571 & 0.647 & \textbf{0.653} \\
7 & 0.613 & 0.556 & 0.638 & \textbf{0.642} \\
\hline
\end{tabular}
\caption{Exact Match average across number of digits.}
\label{tab:exact_match_digits}
\end{table*}

For a more granular observation, we include in Appendix \ref{sec:app_results_agg_tasks} results reporting Exact Match accuracy aggregated by task group. 

\section{Discussion}

The results indicate that our proposed approach consistently outperforms the standard Transformer baseline across all evaluation metrics on our subset of the NUPA benchmark. In particular, we observe an improvement of over 3 percentage points in Exact Match, suggesting that explicitly modeling numerical value yields substantial benefits for precise numerical reasoning. This performance gap is consistent across tasks and digit lengths, indicating that the gains are not driven by isolated cases.

In our approach, we use and predict the magnitudes of numerical values, similar to Numerologic \citep{schwartz2024numerologic}. However, a key limitation of Numerologic is that, despite reasoning about magnitude, it still represents magnitude as a discrete token. As a result, the numerical value itself is not encoded; instead, it is mapped to an arbitrary symbol, suffering from the same fundamental issue as standard token-based number representations. Consequently, Numerologic continues to treat numbers as tokens rather than as values. In practice, this means that while the method resembles a form of chain-of-thought reasoning over numerical categories, it does not resolve the core problem of grounding numerical reasoning in the actual quantitative value of numbers.

When comparing the two variants of the \texttt{<num>} embedding module, namely the fixed-length MLP-based encoder and the recurrent (RNN-based) encoder, the results reveal complementary strengths. The RNN-based approach achieves slightly higher Exact Match scores, which may be attributed to its capability to naturally process variable-length digit sequences and to generalize more effectively across different numerical lengths. In contrast, the MLP-based variant shows marginally better performance on the Digits Match and dLength metrics, suggesting improved local digit alignment and length estimation under fixed-size representations.

Despite these differences, the observed performance gaps between the MLP and RNN variants are relatively small, and the results do not support a definitive conclusion regarding the overall superiority of one approach over the other. Nevertheless, we recommend the RNN-based embedding as the default configuration. This recommendation is motivated by its stronger Exact Match performance, its architectural flexibility with respect to the number of digits, without requiring hard-coded maximum lengths as in the fixed-input MLP, and its closer alignment with how numerical sequences are typically processed in human cognition, where numbers are handled as ordered digit sequences rather than fixed-size vectors.

\section{Conclusions and Future Work}

This work addresses a fundamental limitation of Transformer-based language models: the absence of an explicit representation of numerical value. We introduced a value-aware numerical encoding mechanism that augments standard tokenized representations with a dedicated \texttt{<num>} prefix token whose embedding is explicitly conditioned on the following numerical value. This design injects magnitude information directly into the model's input space while remaining compatible with existing tokenizers, architectures, and autoregressive decoding. The approach enables the model to condition numeric generation on a continuous value representation.

An empirical evaluation on the NUPA benchmark shows that this explicit value modeling leads to consistent improvements over both a standard Transformer baseline and a strong text-based magnitude-aware method. Gains are observed across numerical formats, task types, and operand lengths, with particularly robust behavior as numerical magnitude increases. Beyond the specific architecture explored here, the proposed mechanism highlights a broader modeling principle: numerical values should be treated as signals rather than as emergent properties of token co-occurrence. 

Several directions remain open for future work. A natural extension is to evaluate the proposed value-aware representation in conjunction with large pretrained language models across a broader range of reasoning-intensive tasks and mathematical problem benchmarks to assess the additional benefit that explicit value encoding provides. Furthermore, the design space of the numerical embedding module itself warrants deeper exploration, including alternative architectures, richer auxiliary features, and tighter integration with the Transformer layers. Such analyses would help clarify the conditions under which explicit value modeling is most effective and how it can be scaled to more general-purpose language models.

\section*{Limitations}
The experimental evaluation in this work is conducted at a prototype level and is primarily intended to validate the feasibility and effectiveness of value-aware numerical representations under controlled conditions. To ensure consistency and comparability across methods, all models share the same architecture and hyperparameters; no dedicated hyperparameter tuning was performed, which may impact absolute performance but does not alter the relative comparisons reported.

In addition, experiments are limited to a constrained evaluation setting using models trained from scratch and a benchmark designed to isolate numerical competence. While this setting supports clear analysis of numerical behavior, the results may not fully reflect performance in more complex scenarios involving large pretrained models, reasoning-intensive tasks, or instruction-following settings.

Finally, the method comes with a higher training cost, as it requires two forward passes for each training example. However, this cost doesn't apply during inference, when tokens are generated sequentially anyway.
\bibliography{custom}
\newpage
\appendix

\section{Dataset Example Tasks}
\label{sec:app_dataset_example_tasks}

Table \ref{tab:sample_tasks} showcases a variety of sample tasks drawn from the NUPA benchmark \citep{yang2024numbercookbook}, illustrating the types of numerical operations included in the training and testing setup.

\begin{table*}[t!]
\centering
\small
\begin{tabular}{|>{\raggedright\arraybackslash}p{10cm}|>{\raggedright\arraybackslash}p{2cm}|}
\hline
\textbf{Task} & \textbf{Result} \\
\hline
Add two numbers: 830 + 70 & 900 \\\hline
Add two numbers: 1.7 + 87.954 & 89.654 \\\hline
Add two numbers: 4/3 + 7/6 & 5/2 \\\hline
Add two numbers: 9.65e42 + 7.594e42 & 1.7244e43 \\\hline
Subtract two numbers: 693 - 647 & 46 \\\hline
Subtract two numbers: 16.513 - 1.28 & 15.233 \\\hline
Subtract two numbers: 8/9 - 6/7 & 2/63 \\\hline
Subtract two numbers: 9.711e38 - 9.2e36 & 9.619e38 \\\hline
Multiply two numbers: 211 * 314 & 66254 \\\hline
Multiply two numbers: 4.3 * 454.416 & 1953.9888 \\\hline
Multiply two numbers: 8/5 * 7/6 & 28/15 \\\hline
Multiply two numbers: 3.4e44 * 5.917e95 & 2.01178e140 \\\hline
Get the maximal number: 67530 and 65941 & 67530 \\\hline
Get the maximal number: 1.114 and 1.12 & 1.12 \\\hline
Get the maximal number: 5/6 and 3/5 & 5/6 \\\hline
Get the maximal number: 2.621e28 and 4.9e62 & 4.9e62 \\\hline
The total number of digits of 94789 & 5 \\\hline
The total number of digits of 7.643 & 4 \\\hline
Divide two numbers and return the result as a fraction. 820 / 860 & 41/43 \\\hline
Divide two numbers and return the result as a fraction. (6/5) / (8/9) & 27/20 \\\hline
Divide two numbers and return the result as an integer. 912 // 3 & 304 \\\hline
Divide two numbers and return the remainder. 617 \% 193 & 38 \\\hline
Convert the number to float: 9.864e3 & 9864.0 \\\hline
Convert the number to scientific notation: 59.506 & 5.9506e1 \\\hline
60939 count the occurrence time of digit 9 & 2 \\\hline
Convert the number to scientific notation: 54.213 and keep significant figures as 3 & 5.42e1 \\\hline
\end{tabular}
\caption{Sample tasks from the NUPA benchmark \citep{yang2024numbercookbook}.}
\label{tab:sample_tasks}
\end{table*}

\section{Dataset and Benchmark Details}
\label{sec:app_dataset}

Table \ref{tab:dataset_num_samples} reports the number of samples for each dataset partition (training, validation, and test) used in our experiments. The same data splits and sample counts are employed consistently across all evaluated models, including both the baselines and our proposed approach, to ensure a fair comparison.

All partitions are derived and defined by the NUPA benchmark \citep{yang2024numbercookbook}. For each partition, we retain the majority of task types and exclude only three categories that primarily require digit-level manipulation rather than numerical value understanding: identifying the maximum or minimum digit, extracting a digit at a specific position, and digit-by-digit addition. All remaining task types are preserved, including arithmetic operations (addition, subtraction, multiplication, division, and modulo), comparison tasks (maximum and minimum), representation conversions, counting tasks, and related value-based numerical operations, for all four types of data (integer, float, fraction, and scientific notation).

\begin{table}[t!]
\centering
\small
\begin{tabular}{|p{3.5cm}|p{3cm}|}
\hline
\textbf{Train - by number of digits} & \textbf{Number of samples} \\
\hline
up to 3 & 5,586,474\\
from 3 to 5 & 7,962,412\\
from 5 to 7 & 8,340,750\\
\hline
Total train size & 21,889,636\\
\hline
\hline
\textbf{Validation - by number of digits} & \textbf{Number of samples} \\
\hline
up to 3 & 44,890\\
from 3 to 5 & 69,918\\
from 5 to 7 & 75,904\\
\hline
Total val size & 190,712\\
\hline
\hline
\hline
\textbf{Test} - number of tasks & 38\\
\textbf{Test} - total number of samples & 191,698\\
\hline
\end{tabular}
\caption{Number of samples for dataset partitions.}
\label{tab:dataset_num_samples}
\end{table}

\section{Hyperparameters and Experimental Details}
\label{sec:app_hyperparameters}

Table \ref{tab:hyperparameters} summarizes the hyperparameters and experimental settings used in our study. Hyperparameters listed before the demarcation line are shared across all models, including the baselines and our proposed approach, while those appearing after the line are specific to our method.

\begin{table}[t!]
\centering
\small
\begin{tabular}{|p{2.5cm}|p{4cm}|}
\hline
\textbf{Name} & \textbf{Value} \\
\hline
\hline
\hline
Model used & GPT-2 \footnote{\url{https://huggingface.co/openai-community/gpt2}} \\\hline
Batch size & 256 \\\hline
Optimizer & AdamW \citep{loshchilov2017adamw} \\\hline
Learning rate & 5e-5 \\\hline
Scheduler & 30\% per epoch warmap steps, then cosine decay up to 6e-5, then flat. Restarts on each epoch. \\\hline
Epochs & 1 for each digit size (up to 3, up to 5, up to 7) \\\hline
\hline
\hline
Separate learning rate & 1e-3 for the \texttt{<num>} embedding module \\\hline
Activation function & tanh for the \texttt{<num>} embedding module \\\hline
MLP fixed input size & 18 for digits, 22 for decimals for the MLP \texttt{<num>} embedding module \\\hline
Projection Module & LlamaMLP with tanh as activation \\\hline
Reconstruction loss objective & cosine similarity between embedding and projection \\\hline
Reconstruction loss coefficient & $\lambda = 0.5$ \\\hline
\end{tabular}
\caption{Hyperparameters and experiments details.}
\label{tab:hyperparameters}
\end{table}

\section{Results Aggregated by Tasks}
\label{sec:app_results_agg_tasks}

Table~\ref{tab:exact_match_task_group} reports Exact Match accuracy aggregated by task group. This view isolates differences in model performance across distinct numerical operations and provides insight into how value-aware representations affect various arithmetic and structural tasks.

\onecolumn

\begin{table*}[ht]
\centering
\small
\begin{tabular}{r|cccc}
\hline
\textbf{Task Group} &
\textbf{Standard Transformer} &
\textbf{Numerologic} &
\textbf{NumValue-MLP (ours)} &
\textbf{NumValue-RNN (ours)} \\
\hline
add & 0.660 & 0.533 & \textbf{0.682} & \textbf{0.682} \\
subtract & 0.772 & 0.563 & \textbf{0.788} & \textbf{0.788} \\
multiply & 0.245 & \textbf{0.545} & 0.340 & 0.367 \\
truediv & 0.464 & 0.312 & 0.635 & \textbf{0.646} \\
floordiv & 0.739 & 0.006 & \textbf{0.760} & 0.709 \\
modulo & \textbf{0.076} & 0.000 & 0.059 & 0.051 \\
count & 0.994 & 0.899 & 0.995 & \textbf{0.996} \\
length & \textbf{1.000} & 0.175 & 0.998 & 0.999 \\
max & 0.994 & 0.991 & \textbf{0.996} & 0.994 \\
sig & 0.999 & \textbf{1.000} & \textbf{1.000} & \textbf{1.000} \\
to & \textbf{0.996} & 0.991 & \textbf{0.996} & 0.993 \\
\hline
\end{tabular}
\caption{Exact Match average across task group.}
\label{tab:exact_match_task_group}
\end{table*}

\twocolumn

\end{document}